# Multi-objective semi-supervised clustering to identify health service patterns for injured patients

Hadi Akbarzadeh Khorshidi, Uwe Aickelin, Gholamreza Haffari, Behrooz Hassani-Mahmooei

**Abstract**

**Purpose:** This study develops a pattern recognition method that identifies patterns based on their similarity and their association with the outcome of interest. The practical purpose of developing this pattern recognition method is to group patients, who are injured in transport accidents, in the early stages post-injury. This grouping is based on distinctive patterns in health service use within the first week post-injury. The groups also provide predictive information towards the total cost of medication process. As a result, the group of patients who have undesirable outcomes are identified as early as possible based health service use patterns.

**Methods:** We propose a multi-objective optimization model to group patients. An objective function is the cost function of *k*-medians clustering to recognize the similar patterns. Another objective function is the cross-validated root-mean-square error to examine the association with the total cost. The best grouping is obtained by minimizing both objective functions. As a result, the multi-objective optimization model is a semi-supervised clustering which learns health service use patterns in both unsupervised and supervised ways. We also introduce an evolutionary computation approach includes stochastic gradient descent and Pareto optimal solutions to find the optimal solution. In addition, we use the decision tree method to reproduce the optimal groups using an interpretable classification model.

**Results:** The results show that the proposed multi-objective semi-supervised clustering identifies distinct groups of health service use s and contributes to predict the total cost. The performance of the multi-objective model has been examined using two metrics such as the average silhouette width and the cross-validation error. The examination proves that the multi-objective model outperforms the single-objective ones. In addition, the interpretable classification model shows that imaging and therapeutic services are critical services in the first-week post-injury to group injured patients.

**Conclusion:** The proposed multi-objective semi-supervised clustering finds the optimal clusters that not only are well-separated from each other but can provide informative insights regarding the outcome of interest. It also overcomes two drawback of clustering methods such as being sensitive to the initial cluster centers and need for specifying the number of clusters.

## Introduction

Transport accidents play an important role in public health as they are a leading cause for injury and mortality globally [1]. Therefore, an early distinct grouping of patients who are injured in transport accidents is informative for providing a customized level of care for patients [2]. This study aims to group injured patients based on minimal historical information in such a way that groups are meaningfully related to the outcome of interest. The historical information, we investigate in this study, is the health service used by patients in the first-week post-injury. The health service use is a rich source of information that can provide insights for the recovery process [3, 4], however, to the best of our knowledge, grouping patients based on health service use patterns and their association with injury outcomes has not been studied. In this study, we considered the total cost of recovery as the outcome of interest. This study gives insights to health service providers and health insurance policy makers for making appropriate decision by identifying the patients, with a high-risk of costly medication, as early as possible.

In this study, we use the payment data for patients who are injured within the state of Victoria, Australia. This data is part of a bigger cohort of data which recorded by the Transport Accident Commission and stored in an administrative database called the Compensation Research Database [5]. The sample data includes the patients with a non-fatal transport injury who their injury had occurred between 1 January 2010 and 31 December 2012 and their age at time of injury was over 15 and under 80 years old. After cleaning the data, the sample includes more than 1.6 million payment records which belong to 20,692 unique patients. We then transformed the dataset in a way to be used to achieve the study's aims. As a result, in the transformed data, each row represents a patient, each column is a health service and the values are the amount paid on each health service for a patient within the first week post-injury. There are 19 columns (health services) in the dataset. These services are categorized in eight standard Medicare categories [6] as ambulance services, pharmacy, hospital services, medical services, paramedical services, long-term care, vocational and educational and other services include prosthetics and legislated post-hospital support.

We developed a semi-supervised clustering method using multi-objective optimization to achieve the aims of this study. Both unsupervised and supervised information are used in semi-supervised clustering [7]. Semi-supervised clustering has attracted much interest recently [8-10]. Our proposed semi-supervised clustering optimally divides the patients in groups based on both health service use and association with the total cost simultaneously. The proposed multi-objective optimization model is to find the optimal clusters (groups) that minimize both the clustering cost function and the prediction error. To find the optimal groups, i.e. solve the optimization model, we introduce an evolutionary computation approach which uses stochastic gradient descent to update solution across generations. Also, as we solve a multi-objective optimization model, the Pareto optimal (non-dominated) solutions are considered.

The hypothesis for developing a multi-objective optimization model is that it works better than single-objective ones to achieve the aims of the study as it intends to contribute to both objective functions. In addition, reformulating a multi-objective optimization model into a weighted average objective function or a goal programming model leads to a failure for finding non-dominated optimal solutions [11].

The main research question in this study is to recognize patterns of utilizing health services for injured patients based on the similarity of the patterns and their association with the outcome of interest. Identifying these patterns as early as possible provides insights about the medication procedure of the patients. Developing a semi-supervised clustering is motivated by taking advantages of multi-objective optimization models. The proposed multi-objective semi-supervised clustering also contributes to overcome two drawback of unsupervised clustering methods as being sensitive to the initial solutions and dependency on specifying the number of clusters in advance.

The proposed semi-supervised clustering is implemented on the transformed dataset of transport accident injured patients. The optimal groups of patients are found based on the health service use in the first-week post-injury through the implementation. These groups are well-separated and informative about the total cost of medication. Once the optimal groups are found, we compare the performance of the multi-objective optimization model with single-objective ones. At the end, we use the decision tree method to develop a set of operational rules to identify the optimal groups. it should be noted that this paper is an extended version of [12].

## Multi-objective semi-supervised clustering
### Objective functions

An objective function which we use in the proposed optimization model is the cost function of the $k$-medians clustering. The $k$-medians clustering is a robust clustering method and is less sensitive to outliers in comparison with $k$-means clustering [13]. In $k$-medians clustering, the objective is to minimize the function formulated in Eq. 1.

$$f(X) = \sum_{r=1}^{k} \mathbb{E}[I_r(Z;X)\|Z - X^r\|_1] \quad (1)$$

where $Z \in \mathbb{R}^d$, $X = (X^1, \dots, X^k)'$ and $X^r$ is the center of $r$th cluster, $\|.\|_1$ is the norm 1 and calculates the absolute distance, and $I_r$ is the indicator as defined by Eq. 2.

$$I_r(z;X) = \prod_{j=1}^{k} \mathbb{1}_{\{\|z-X^r\|_1 \leq \|z-X^j\|_1\}} \quad (2)$$

which means that $X^r$ is the nearest center to $z$. So, data points are assigned to the cluster with the nearest centre.

Another objective function aims to measure how much the clusters are associated with the outcome of interest. It estimates the cross-validated error of prediction for the multivariate linear regression model where clusters are predictors. Therefore, in the regression model, $I$ is the matrix of predictors which is constructed based on $X$ (the set of cluster centres) so that $I_{ik}$ is one if $i$th record is allocated to $k$th cluster; otherwise it is zero. So, the second objective function minimizes the $m$-fold Cross-validation estimation of root-mean-square error as formulated in Eq. 3.

$$g(X) = \frac{1}{M}\sum_{m=1}^{M} \sqrt{\frac{\sum_{i \in F_m}(y_i - \hat{y}_i)^2}{n_m}} \quad (3)$$

where $M$ is the total number of folds and $F_m$ is the $m$th fold in the dataset, $n_m$ is the number of records in $F_m$, $y_i$ denotes the actual values of outcome and $\hat{y}_i$ represents the predicted values which are dependent on the set of cluster centres. The $\hat{Y}$, the set of predicted values for the outcome of interest, can be calculated by Eq. 4.

$$\hat{Y} = [\hat{y}_i]_{n_m \times 1} = [1, I]_{n_m \times (k+1)} \beta_{(k+1) \times 1} \quad (4)$$

where $\beta$ is estimated based on cluster assignment ($I = [I_{ik}]$) and the set of actual values of the outcome ($Y = [y_i]$) when $i$ is not a member of $F_m$ (as formulated in Eq. 5).

$$\beta = ([1,I]'[1,I])^{-1}[1,I]'Y \quad (5)$$

The proposed semi-supervised clustering problem can be completed as the bi-objective optimization model is formulated in Eq. 6 at the end.

$$\begin{cases} Min\ f(X) \\ Min\ g(X) \end{cases} \quad (6)$$

To calculate the objective functions, we need to create distance and allocation matrices. The function to create these matrices is outlined in Algorithm 1.

**Algorithm 1:** Allocation-Distance function

**Input:** A dataset $Z$ and a set of cluster centres $X$
**Output:** A distance matrix $D$ and an allocation matrix $I$

1. $n \leftarrow$ number of rows of $Z$; $k \leftarrow$ number of rows of $X$
2. $D \leftarrow$ An empty matrix with $n$ rows and $k$ columns
3. $I \leftarrow$ An empty matrix with $n$ rows and $k$ columns
4. **for** $i$ from 1 to $n$ **do**
5.   **for** $j$ from 1 to $k$ **do**
6.     $D_{ij} = sum(|Z_i - X^j|)$
7.   **end for**
8.   **for** $j$ from 1 to $k$ **do**
9.     **if** $D_{ij} = \min_i D_{ij}$ **then** $I_{ij} = 1$ **else** $I_{ij} = 0$
10.   **end for**
11. **end for**

## Optimal solutions

After constructing the multi-objective semi-supervised clustering model, an algorithm is needed to solve the optimization problem in order to find the optimal solutions. In this section, we introduce an evolutionary algorithm which reach to the optimal set of cluster centres iteratively. To update the set of cluster centres across iterations, we use stochastic gradient for $k$-medians clustering which is proposed in [14]. The cluster centres are updated using Eq. 7 in the next iteration.

$$X^r_{h+1} = X^r_h - a^r_h I_r(z;X) \frac{x^r_h - z}{\|x^r_h - z\|_1} \quad (7)$$

where $X^r_h$ is the center of $r$th cluster in $h$th iteration and $a^r_h$ is the learning rate which is calculated by Eq. 8.

$$a^r_h = \frac{c_\gamma}{(1+c_\alpha n_r)^\alpha} \quad (8)$$

where $n_r$ is the number of data points in the $r$th cluster and $\alpha$, $c_\gamma$ and $c_\alpha$ are parameters with constant values.

A drawback for $k$-medians clustering (also for $k$-means clustering) is that it is sensitive to the choice of the initial cluster centres [14-16] i.e. different initial centres lead to different clusters at the end of algorithm. To overcome this drawback, we initialize the iteration process with a pool of the sets of cluster centres.

Another drawback is that the number of clusters should be specified at the beginning. However, in many cases of unsupervised learning, the researchers or practitioners are not aware of the number of clusters. Many studies have been conducted to address this issue [17-20]. In our algorithm, we assign different number of clusters for each set in the initial pool of cluster centres. As a result, the optimization model not only finds the best clusters by minimizing both clustering cost function and regression cross-validation error but also determines the best number of clusters.

In each iteration, the algorithm calculates the values of objective functions for all sets of cluster centres. As there are two objective functions, we use Pareto optimal solutions to find the best sets. The Pareto optimal solutions are the solutions that are not dominated by any other solution [21, 22]. In the proposed multi-objective semi-supervised clustering, the Pareto optimal solutions are the sets of cluster centres where there is no other set with lower values in both objective functions as both objective functions are to be minimized. We determine the Pareto optimal sets in each iteration and keep them in the next iteration, while we update the other sets using Eq. 7.

The stopping criterion for the iterative algorithm is set to reach a prespecified maximum number of iterations ($tau_{max}$). The best set of cluster centres is selected among the Pareto optimal solutions in the last iteration. We firstly normalize the objective function values of the Pareto optimal sets using Eq. 9 [23].

$$NormPOS = \frac{POS_{po} - \overline{POS_o}}{sd(POS_o)} | p = 1,2,\dots,n_p; o = 1,2 \quad (9)$$

where $POS_{po}$ refers to the $o$th objective function value for the $p$th Pareto optimal solution, $n_p$ is the number of Pareto optimal solutions in the last iteration, $\overline{POS_o}$ and $sd(POS_o)$ are the mean and standard deviation values of the $o$th objective function for Pareto optimal solutions.

Then, we determine the positive ideal solution which includes the minimum value of objective functions across all Pareto optimal solutions (formulated as Eq. 10). After that, we measure the similarity between each Pareto optimal solution and the positive ideal solution using the cosine similarity measure as Eq. 11 [24, 25]. Finally, we select the Pareto optimal solution with the highest similarity value as the best set of cluster centres which denotes the optimal clusters.

$$PI = \left(\min_o NormPOS_{po} | p = 1,2,\dots,n_p; o = 1,2\right) = (PI_1, PI_2) \quad (10)$$

$$Similarity(NormPOS_p, PI) = \cos(\theta) = \frac{NormPOS_p \cdot PI}{\|NormPOS_p\|_2 \cdot \|PI\|_2} \quad (11)$$

where $PI$ is the positive ideal vector, $\|.\|_2$ is the norm 2 to calculate the magnitude of each vector and $NormPOS_p \cdot PI$ is the inner product of two vectors.

The proposed evolutionary algorithm is outlined in Algorithm 2.

**Algorithm 2:** The proposed semi-supervised clustering

**Input:** A dataset $Z$; A vector for the outcome of interest $Y$
    The maximum number of clusters $K$
    Number of sets in the pool $n_{pool}$
    Maximum number of iterations ($Iter_{max}$) /* Stopping criterion */
    Learning rate parameters $\alpha$, $c_\gamma$ and $c_\alpha$
**Output:** The optimal set of cluster centres x

1. **for** $r$ from 1 to $n_{pool}$ **do**
2.   **for** $k$ from 2 to $K$ **do**
3.     $Xr \leftarrow$ Randomly select $k$ samples from $Z$ as a set of cluster centres

4.  **end for**
5. **end for**
6. $ObjFunc \leftarrow$ An empty matrix with $Iter_{max} \times n_{pool}$ rows and 2 columns
7. $tau \leftarrow 1$
8. **while** $tau \leq tau_{max}$ **do**
9.  **for** $r$ from 1 to $n_{pool}$ **do**
10.   $Dr \leftarrow$ Distance matrix via applying algorithm 1 on $Z$ and $Xr$
11.   $Ir \leftarrow$ Allocation matrix via applying algorithm 1 on $Z$ and $Xr$
12.   $ObjFunc[,1] \leftarrow f(Xr)$ /* Using Eq. 1 */
13.   $ObjFunc[,2] \leftarrow g(Xr)$ /* Using Eq. 3 */
14.  **end for**
15.  $n \leftarrow 0; P \leftarrow$ An empty vector to record Pareto solutions
16.  **for** $r$ from 1 to $n_{pool}$ **do**
17.   $m \leftarrow 0$
18.   **for** $t$ from 1 to $n_{pool}$ not including $r$ **do**
19.    **if** both $ObjFunc$ values of set $r$ are bigger than set $t$ **then** $m = m + 1$
20.   **if** $m > 0$ **then**
21.    **for** $j$ from 1 to the number of rows of $Xr$ **do**
22.     $Zq \leftarrow$ Randomly select a sample from data points allocate to cluster $j$
23.     Update the cluster centre $j$ by $Zq$ /* Using Eq. 6 */
24.    **end for**
25.   **else**
26.    $n = n + 1$
27.    $P[n] = r$
28.   **end if**
29.  **end for**
30.  $tau = tau + 1$
31. **end while**
32. $NormP \leftarrow$ Normalize the objective values of Pareto solutions /* Using Eq. 8 */
33. $PI \leftarrow$ Find the positive ideal solution /* Using Eq. 10 */
34. **for** $l$ from $P[1]$ to $P[n]$ **do**
35.  $S[l] \leftarrow$ Calculate similarity between $PI$ and $NormP[l,]$ /* Using Eq. 11 */
36. **end for**
37. The set with the maximum value in $S$ is the optimal set

## Implementation

In this section, we present the results of implementing the proposed semi-supervised clustering on the transformed dataset of the first-week post-injury health service use of injured patients. The outcome of interest is the total cost to recovery. The aim of this implementation is to find the separate groups of patients who have similar patterns of health service use and also have distinct level of the total costs at the end of their medication process. As a result, the group of patients who may lead to costly medication process (unwanted outcomes) can be identified at the early stages of their medication for receiving extra care.

For this implementation, we specify the number of cluster centre sets in the initial pool as 100 ($n_{pool}$=100). The number of clusters varies from 2 to 11 ($K$=11). The iteration process continues for 500 times ($tau_{max}$=500). For the learning rate, we set parameters as $\alpha = 3/4$, $c_\alpha = 1$ (as recommended in [14]), and $c_\gamma = 2000$ (based on experiments).

Fig. 1 (a) shows the initial pool of cluster centre sets using their clustering's cost function and regression's cross-validation error values. The red points are the Pareto optimal solutions. Fig. 1 (b) visualizes the final pool of cluster centre sets once the algorithm stops after 500 iterations. As can be seen, the number of Pareto optimal solutions increases, and sets move towards the Pareto frontier during the iteration process.

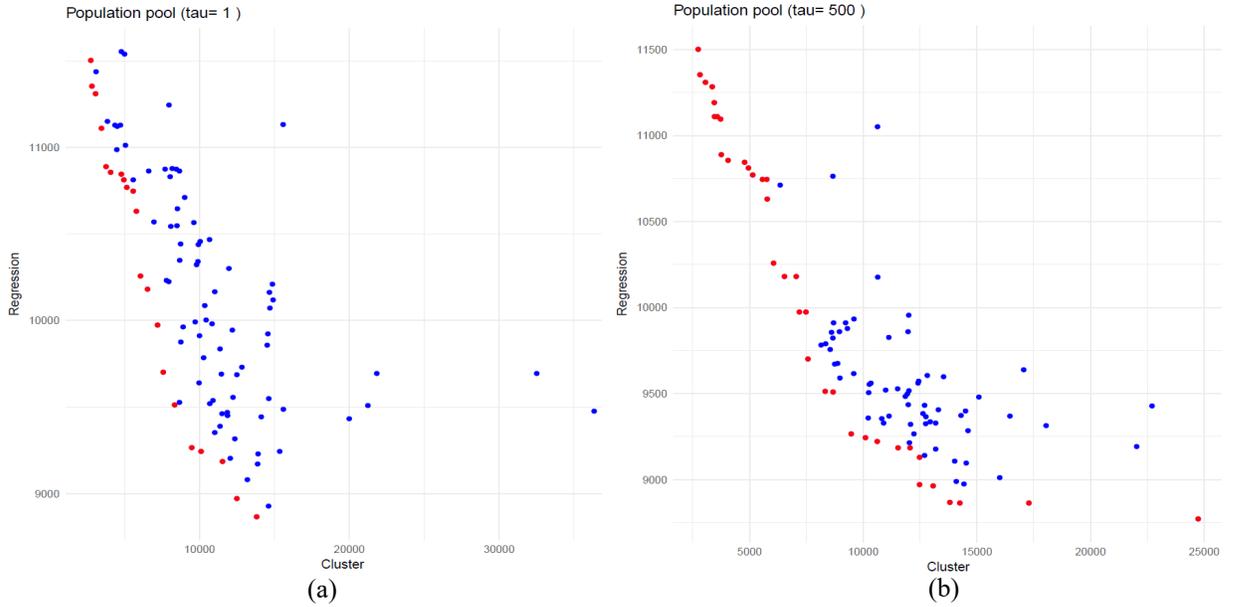

**Fig. 1** The pool of sets in the first iteration (a) and the 500[th] iteration (b)

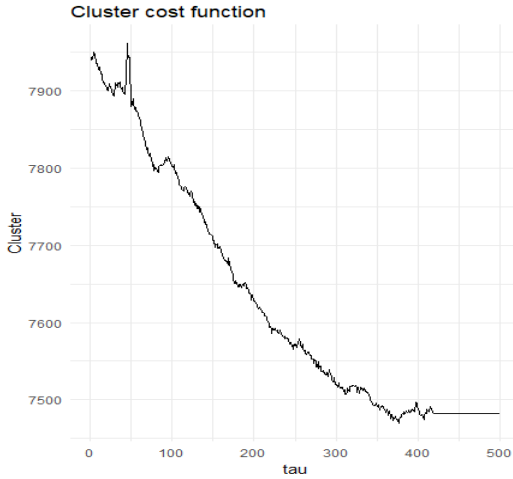
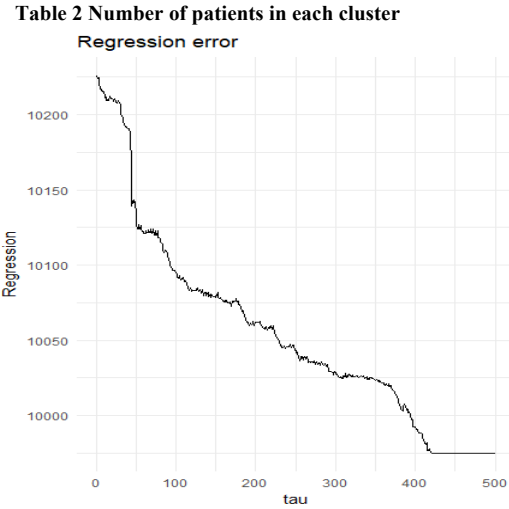

**Fig. 2** Cluster cost function (left) and regression CV error (right) trajectories

There are 36 Pareto optimal solutions in the final pool. Table 1 gives a summary on the frequency of number of clusters among Pareto optimal solution sets.

**Table 1** Frequency of number of clusters

| Number of clusters | 2 | 3 | 4 | 5 | 6 | 7 | 8 | 9 | 10 | 11 |
|---|---|---|---|---|---|---|---|---|---|---|
| Frequency | 10 | 10 | 6 | 4 | 0 | 1 | 2 | 1 | 1 | 1 |

We find the optimal set among Pareto optimal solutions based on their similarity with the positive ideal solution as described in Algorithm 2. The final optimal set contains three cluster centres i.e. it divides the patients into three groups. Fig. 2 shows different values of objective functions for the optimal set during the iterative algorithm. The optimal set becomes a Pareto optimal solution after 415 iterations.

Fig. 3 visualizes the clusters resulted from the optimal set of cluster centres using the first principal component (PC1) and the second principal component (PC2). It shows how well these clusters are separated from each other. Table 2 shows how many patients has been allocated into each cluster.

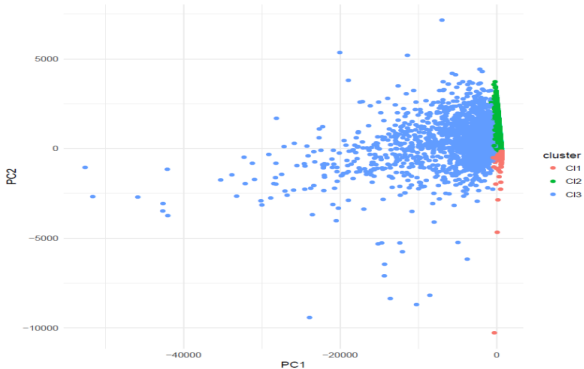

**Fig. 3** Cluster visualization

**Table 2** Number of patients in each cluster

| Cluster | Cluster 1 | Cluster 2 | Cluster 3 |
|---|---|---|---|
| Number of patients | 12,206 | 6,396 | 2,090 |

To compare the performance of the proposed multi-objective optimization model against the single-objective ones, we find the best sets of cluster centres when we just have one of the objective functions. For this comparison, we use two metrics. The first metric is the average silhouette width [18, 23, 26], which is a validity index, to measure the clustering performance. The average silhouette width value lies between -1 and 1 and the higher value demonstrates a better partitioning result. The second one measures the prediction accuracy using the cross-validation error of the regression model which the logarithm of total cost is the response variables and clusters are predictors. Table 3 presents the comparison results using the metric values of the optimal solutions, the average and standard deviation of the final pool. The comparison results show that the multi-objective optimization model provides clusters with better average silhouette width values. The results show that the single-objective model with the objective function of regression error find the optimal set with the lowest cross-validation error. However, the average silhouette width value for this set is very low. In addition, the average value of cross-validation error in the final pool for the multi-objective optimization model is lower than the single-objective ones. As a result, the multi-objective semi-supervised clustering model outperforms two other models by considering both metrics.

**Table 3** Comparison results for optimization models

| Model | Multi-objective | | k-Medians | | Regression | |
|---|---|---|---|---|---|---|
| Metric | average silhouette width | cross-validation | average silhouette width | cross-validation | average silhouette width | cross-validation |
| Optimal solution | 0.505 | 1.239 | 0.109 | 1.527 | -0.019 | 1.199 |
| Average | 0.109 | 1.247 | 0.019 | 1.359 | -0.011 | 1.266 |
| standard deviation | 0.246 | 0.095 | 0.267 | 0.091 | 0.236 | 0.106 |

The implementation results show that the optimal clusters can provide informative insights regarding the total cost of medication. However, we need an interpretable algorithm to allocate future patients into the right groups based on the first-week post-injury health service use. To reach this goal, we use decision tree method to develop the algorithm. The developed algorithm can be interpreted into operational procedures that can be used easily by operators to find the proper groups for patients.

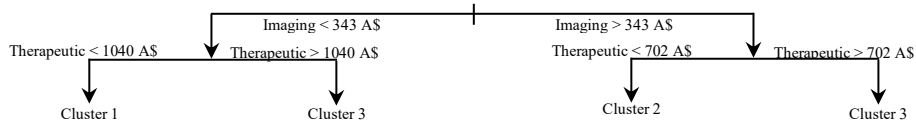

**Fig. 4** Decision tree

Fig. 4 shows how the decision tree allocates the patients into resultant clusters by defining a set of rules. The classification and regression tree method is used to create the decision tree. This classification model provides a transparent algorithm to group injured patients at the early stages. As can be seen, imaging and therapeutic services play an important role to define the classification rules.

We also investigate the performance of the interpretable algorithm deriven by decision tree in predicting the resulted clusters. We set half of the dataset as training and the rest as testing datasets. We compare the testing dataset's misclassification rate of the algorithm with classifaction models created by linear discriminant analysis (LDA) and naïve Bayes classifiers. Table 4 shows the comparison results where we can see the algorithm has the lowest misclassification rate. So, decision tree not only provides an interpretable algorithm, but also classify patients into their true clusters accurately.

**Table 4 Classification comparison results**

| Model | Decision tree | LDA | Naïve Bayes |
|---|---|---|---|
| **Misclassification rate (%)** | 2.1 | 13.8 | 31.4 |

## Conclusions

In this paper, we propose a semi-supervised clustering approach to group transport accident injured patients based on using similar patterns of health services in the first week post-injury and having similar total cost. This grouping is helpful for health service providers and policy makers to identify groups which need special care in the early stages of medication process using the minimum historical information.

For semi-supervised clustering, we develop a multi-objective optimization model with two objective functions for clustering and regression. We use stochastic gradient descent and Pareto optimal solutions within an evolutionary algorithm to find the optimal solutions. The proposed semi-supervised clustering not only finds the best clusters by minimizing both clustering cost function and regression cross-validation error but also contributes to overcome two drawback of clustering methods such as sensitivity to the initial cluster centres and prespecifying the number of clusters.

The implementation of the proposed semi-supervised method shows that the resultant clusters identify distinct groups of health service uses and contribute to predict the total cost.

Also, we examine the performance of the multi-objective model using average silhouette width and cross-validation metrics. The examination proves that the research hypothesis is true, and the multi-objective approach outperforms the single-objective approaches. Finally, we use decision tree method as an interpretable algorithm to classify patients to clusters. The classification algorithm denotes that using imaging and therapeutic services in the first-week post-injury are critical services in health service use patterns.

For further research, we suggest implementing the proposed semi-supervised learning in different medical informatics problems. In addition, the various injury outcomes such return to work (RTW) [27] can be considered.

## Acknowledgement

This project was funded by the Transport Accident Commission (Transport Accident Commission) through the Institute for Safety, Compensation and Recovery Research (ISCRR).

## Compliance with Ethical Standards

### Conflict of interest
All authors declare that they have no conflict of interest.
### Ethical Approval
This article does not contain any studies with human participants performed by any of the authors.
### Informed Consent
Statement not required. This study was performed using a de-identified administrative dataset, with ethics approval granted by Monash University Human Research Ethics Committee (CF09/3150—2009001727).

## REFERENCES

[1] Azmin M, Jafari A, Rezaei N, Bhalla K, Bose D, Shahraz S, Dehghani M, Niloofar P, Fatholahi S, Hedayati J, Jamshidi H, and Farzadfar F. An Approach Towards Reducing Road Traffic Injuries and Improving Public Health Through Big Data Telematics: A Randomised Controlled Trial Protocol. *Archives of Iranian medicine*. 2018;21 (11):495-501.

[2] Scheetz LJ, Zhang J, and Kolassa J. Classification tree modeling to identify severe and moderate vehicular injuries in young and middle-aged adults. *Artificial Intelligence in Medicine*. 2009;45 (1):1-10.

[3] Mitchell RJ, Cameron CM, and McClure R. Patterns of health care use of injured adults: A population-based matched cohort study. *Injury*. 2017;48 (7):1393-1399.

[4] Pinaire J, Azé J, Bringay S, and Landais P. Patient healthcare trajectory. An essential monitoring tool: a systematic review. *Health Information Science and Systems*. 2017;5 (1):1-18.

[5] Prang KH, Hassani-Mahmooei B, and Collie A. Compensation Research Database: population-based injury data for surveillance, linkage and mining. *BMC Research Notes*. 2016;9 (1):1-11.

[6] Department of Health, *MBS online*. 2017: http://www9.health.gov.au/mbs/search.cfm?adv=1.

[7] Saha S, Ekbal A, and Alok AK. Semi-supervised clustering using multiobjective optimization. in *Proceedings of the 2012 12th International Conference on Hybrid Intelligent Systems, HIS 2012*. 2012.

[8] Handl J and Knowles J. On semi-supervised clustering via multiobjective optimization. in *GECCO 2006 - Genetic and Evolutionary Computation Conference*. 2006.

[9] Santos L, Veras R, Aires K, Britto L, and Machado V. Medical Image Segmentation Using Seeded Fuzzy C-means: A Semi-supervised Clustering Algorithm. in *Proceedings of the International Joint Conference on Neural Networks*. 2018.

[10] Yang J, Sun L, and Wu Q. Constraint projections for semi-supervised spectral clustering ensemble. *Concurrency Computation*. 2019.

[11] Charkhgard H and Eshragh A. A new approach to select the best subset of predictors in linear regression modelling: bi-objective mixed integer linear programming. *ANZIAM Journal*. 2019;61 (1):64-75.

[12] Khorshidi HA, Haffari G, Aickelin U, and Hassani-Mahmooei B, Early identification of undesirable outcomes for transport accident injured patients using semi-supervised clustering, in *Health Informatics Conference* 2019: Melbourne, Australia.

[13] García-Escudero LA, Gordaliza A, Matrán C, and Mayo-Iscar A. A review of robust clustering methods. *Advances in Data Analysis and Classification*. 2010;4 (2):89-109.

[14] Cardot H, Cénac P, and Monnez JM. A fast and recursive algorithm for clustering large datasets with k-medians. *Computational Statistics and Data Analysis*. 2012;56 (6):1434-1449.

[15] Rahim MS and Ahmed T. An initial centroid selection method based on radial and angular coordinates for K-means algorithm. in *20th International Conference of Computer and Information Technology, ICCIT 2017*. 2018.

[16] Khan F. An initial seed selection algorithm for k-means clustering of georeferenced data to improve replicability of cluster assignments for mapping application. *Applied Soft Computing*. 2012;12 (11):3698-3700.

[17] Bezdek JC and Pal NR. Some new indexes of cluster validity. *IEEE Transactions on Systems, Man, and Cybernetics, Part B (Cybernetics)*. 1998;28 (3):301-315.

[18] Campello RJGB and Hruschka ER. A fuzzy extension of the silhouette width criterion for cluster analysis. *Fuzzy Sets and Systems*. 2006;157 (21):2858-2875.

[19] Nikfalazar S, Yeh C-H, Bedingfield S, and Khorshidi HA. A new iterative fuzzy clustering algorithm for multiple imputation of missing data. in *IEEE International Conference on Fuzzy Systems (FUZZ-IEEE)*. 2017. Naples, Italy.

[20] Sun H, Wang S, and Jiang Q. FCM-Based Model Selection Algorithms for Determining the Number of Clusters. *Pattern Recognition*. 2004;37 (10):2027-2037.

[21] Abouei Ardakan M and Rezvan MT. Multi-objective optimization of reliability–redundancy allocation problem with cold-standby strategy using NSGA-II. *Reliability Engineering & System Safety*. 2018; 172 225-238.

[22] Alok AK, Saha S, and Ekbal A. Semi-supervised clustering for gene-expression data in multiobjective optimization framework. *International Journal of Machine Learning and Cybernetics*. 2017;8 (2):421-439.

[23] Milligan GW and Cooper MC. A study of standardization of variables in cluster analysis. *Journal of Classification*. 1988;5 (2):181-204.

[24] Forestier G, Petitjean F, Senin P, Riffaud L, Henaux PL, and Jannin P. Finding discriminative and interpretable patterns in sequences of surgical activities. *Artificial Intelligence in Medicine*. 2017;82 11-19.

[25] Nikfalazar S, Khorshidi HA, and Hamadani AZ. Fuzzy risk analysis by similarity-based multi-criteria approach to classify alternatives. *International Journal of Systems Assurance Engineering and Management*. 2016;7 (3):250-256.

[26] Mihaljević B, Benavides-Piccione R, Guerra L, DeFelipe J, Larrañaga P, and Bielza C. Classifying GABAergic interneurons with semi-supervised projected model-based clustering. *Artificial Intelligence in Medicine*. 2015;65 (1):49-59.

[27] Zhang J, Cao P, Gross DP, and Zaiane OR. On the application of multi-class classification in physical therapy recommendation. *Health Information Science and Systems*. 2013;1 (1):15.